\documentclass[runningheads]{llncs}

\usepackage{eccv}
\usepackage{eccvabbrv}
\usepackage{graphicx}
\usepackage{booktabs}
\usepackage[table]{xcolor}
\definecolor{upgreen}{rgb}{0.0, 0.5, 0.0}
\usepackage[export]{adjustbox}
\usepackage{array}
\usepackage[accsupp]{axessibility}

\usepackage[breaklinks,colorlinks,citecolor=eccvblue,linkcolor=eccvblue,urlcolor=eccvblue]{hyperref}

\hypersetup{
    pdftitle={DeGuNet: Depth-Guided Ultra-Compact Backbones for Efficient LiDAR-Camera 3D Detection},
    pdfauthor={Haifa Zhang, Yijing Wang, Peixi Peng, and Zhiqiang Zuo}
}

\begin{document}

\title{DeGuNet: Depth-Guided Ultra-Compact Backbones for Efficient LiDAR-Camera 3D Detection} 

\titlerunning{DeGuNet for Efficient LiDAR-Camera 3D Detection}

\author{Haifa Zhang\inst{1} \and
Yijing Wang\inst{1}\thanks{Corresponding authors.} \and
Peixi Peng\inst{2} \and
Zhiqiang Zuo\inst{1}\textsuperscript{$\star$}}

\authorrunning{H.~Zhang et al.}

\institute{Tianjin University, Tianjin, China\\
\email{\{zhanghaifa,yjwang,zqzuo\}@tju.edu.cn}
\and
Peking University, Beijing, China\\
\email{pxpeng@pku.edu.cn}}

\maketitle

\begin{abstract}
In autonomous driving perception, the fusion of LiDAR and camera modalities has become the dominant paradigm for 3D object detection. However, current multi-modal frameworks heavily rely on massive visual backbones pretrained on 2D semantic tasks. This reliance introduces substantial parameter redundancy and a structural misalignment, as 2D priors are ill-equipped to handle the extreme sparsity of LiDAR projections required for Bird's-Eye-View geometry. To address this, we present DeGuNet, an ultra-compact and plug-and-play image backbone explicitly designed for depth-guided representation learning. By incorporating sparsity-aware feature extraction mechanisms, DeGuNet effectively aligns multi-view images with unstructured LiDAR depth while strictly preventing invalid-region contamination. Extensive experiments on the nuScenes dataset demonstrate DeGuNet's broad plug-and-play applicability and superior efficiency. When integrated into established baselines, it fundamentally eliminates architectural redundancy, reducing GPU memory consumption by up to 66.5\% and achieving a 1.16$\times$ inference speedup. Concurrently, DeGuNet delivers up to a 6.20 absolute mAP gain, establishing a new paradigm for parameter-efficient multi-modal 3D perception.


\end{abstract}

\section{Introduction}
\label{sec:intro}

In autonomous driving perception, multi-modal 3D object detection~\cite{cai2023objectfusion,song2023graphalign,chen2022autoalignv2,li2022deepfusion} fusing LiDAR and camera data has become the dominant paradigm~\cite{liu2022bevfusion,lei2025bevheight,bai2022transfusion,chen2023futr3d,chen2022autoalignv2}. While LiDAR provides precise geometric structure~\cite{fan2023super,chen2023voxelnext,feng2024towards,gilmer2017neural,qi2017pointnet++}, cameras offer crucial high-resolution textures and semantic details~\cite{li2022bevformer,liu2024ray,zong2023Temporal}. Current state-of-the-art frameworks achieve remarkable performance by projecting features from both modalities into a unified Bird's-Eye-View (BEV) representation. However, this success heavily relies on large-scale visual backbones, such as large ResNets or Transformers, which are pretrained on 2D semantic tasks like image classification or object detection~\cite{liu2021swin,2020wangcsp,he2016deep}. Consequently, the image branch often dominates the system's parameter count. Empirical profiling of prevailing multi-modal frameworks indicates that standard visual backbones consume between 30M and 78M parameters, accounting for 50\% to 86\% of the total model size. This highlights a severe system-level resource redundancy for a modality that primarily serves a complementary role.

Beyond computational overhead~\cite{chen2022autoalignv2,xie2023sparsefusion}, this reliance on parameter-heavy 2D backbones leads to a structural inconsistency. Features optimized for 2D semantic tasks lack the vital 3D spatial awareness required for BEV localization, which inherently depends on accurate metric depth estimation to lift pixels into 3D space. To bridge this gap, a natural intuition is to adopt a depth-centric pretraining paradigm. However, our empirical analysis (detailed in Sec.~\ref{sec:exp}) reveals that simply pretraining standard lightweight backbones (such as a tiny ResNet~\cite{he2016deep} or MobileViT~\cite{mehta2021mobilevit}) on depth completion tasks yields marginal improvements in downstream 3D detection, typically plateauing around 62.1 mAP. We attribute this limited efficacy to the inability of standard convolutional architectures to process extreme spatial sparsity. When projecting sparse LiDAR point clouds onto dense camera planes (over 98\% of the pixels remain empty, as counted in our nuScenes~\cite{caesar2020nuscenes} analysis), standard receptive fields become heavily contaminated by these invalid, zero-value regions. Thus, establishing an effective depth-guided representation requires not only a geometry-centric pretraining objective but also a dedicated, sparsity-aware architecture.

To address these challenges, we propose the Depth-Guided Network (DeGuNet), an ultra-compact, plug-and-play image backbone containing only 0.31M parameters. Instead of inheriting fixed 2D priors, DeGuNet is purposefully pretrained on a depth completion task to generate features intrinsically aligned with the target BEV geometry. To handle sparse geometric guidance, we propose Masked Partial Inverted Residual (MPIR) blocks and progressive Guide modules. These components enforce mask-aware feature extraction to prevent invalid regions from degrading the learned representations. By extracting depth-aware features directly aligned with LiDAR geometry before cross-modal fusion, DeGuNet serves as a general-purpose infrastructure capable of seamlessly replacing heavyweight image backbones across various multi-modal frameworks.

Our main contributions are summarized as follows:
\begin{itemize}
    \item We reveal the critical feature misalignment in current multi-modal 3D detectors and demonstrate that resolving it requires not just depth-centric pretraining, but a specialized sparsity-aware architecture capable of handling extreme sparsity levels over \textbf{98\%} invalid regions.
    \item We propose DeGuNet, a parameter-efficient (0.31M parameters) and plug-and-play image backbone. It integrates MPIR blocks and progressive Guide modules to effectively fuse multi-view images with sparse LiDAR geometry without invalid-region contamination.
    \item Extensive cross-baseline experiments demonstrate DeGuNet's broad plug-and-play applicability. When integrated as a drop-in replacement into established frameworks including BEVFusion, GraphBEV, EA-LSS, and IS-Fusion, it reduces GPU memory consumption by up to \textbf{66.5\%} and achieves a \textbf{1.16$\times$} inference speedup, while concurrently delivering up to \textbf{6.20} absolute mAP gain in 3D detection accuracy.
\end{itemize}

\section{Related Work}
\label{sec:work}

\subsection{Multi-modal 3D Object Detection}
The fusion of LiDAR and camera data has become a widely adopted paradigm for high-performance 3D object detection. 
A prevailing approach projects features from both modalities into a unified BEV representation~\cite{philion2020lift,liu2022bevfusion,liang2022bevfusion}, providing a shared, geometrically consistent grid for cross-modal alignment~\cite{li2022bevformer}. 
Concurrently, query- and transformer-based methods model multi-modal interactions through attention mechanisms~\cite{bai2022transfusion,yan2023cross,yang2022deepinteraction,yang2025deepinteraction++}.
Recently, to address computational bottlenecks in dense BEV maps, several schemes such as SparseFusion~\cite{xie2023sparsefusion} and Fully Sparse Fusion~\cite{li2024fully} have been proposed to operate on sparse, instance-level features.
Despite these advances in the fusion stage, both lines of research remain tethered to bulky 2D-pretrained visual backbones. Their image features are not intrinsically aligned with the geometry-aware BEV space, as BEV localization requires metric depth ordering and cross-view geometric consistency that are not directly encouraged by 2D pretraining.

\subsection{Limitations of Visual Backbones in 3D Perception}
A primary bottleneck exacerbating the aforementioned geometry-semantic misalignment is the conventional visual feature extraction paradigm. Current state-of-the-art multi-modal detectors heavily rely on large-scale visual backbones, such as deep ResNets~\cite{he2016deep} and Swin Transformers~\cite{liu2021swin}, which are pretrained on 2D tasks like ImageNet classification~\cite{ILSVRC15} or 2D object detection. 
While these heavy backbones provide rich semantic priors, they incur considerable parameter redundancy.
Furthermore, the semantic features inherited from fixed 2D pretraining are misaligned with the geometry-aware representations required for 3D prediction, which depend on metric depth cues and cross-view consistency rather than purely appearance semantics. The backbone's original 2D objective is not specifically optimized for 3D supervision, resulting in underutilized feature capacity. 
In contrast, our work discards the conventional large-scale visual backbone. We propose a parameter-efficient architecture that serves as a task-aligned drop-in alternative.

\subsection{Depth-Guided Representation Learning}
Depth information is essential for bridging the modality gap between 2D images and 3D space. 
Many existing 3D detection frameworks incorporate auxiliary depth estimation heads and dense depth losses during detector training (e.g., BEVDepth)~\cite{li2023bevdepth,zhou2023bev} to actively guide the projection process~\cite{li2023lwsis,wang2019pseudo}. 
On a parallel track, depth completion networks~\cite{cheng2019learning,wang2023lrru,yan2022rignet} address the generation of dense depth maps from sparse LiDAR projections. To handle the irregular distribution of sparse points, specialized operations such as partial convolutions~\cite{liu2018partialinpainting} have been explored to apply convolutional filters only to valid pixels.
Our approach differs from standard auxiliary depth strategies used during detector training. Instead of adding depth prediction burdens to the downstream 3D detector, we shift the geometric alignment to a standalone pretraining stage via a depth completion task. By integrating mask-aware partial convolutions~\cite{liu2018partialinpainting} into our MPIR blocks, we strictly exclude invalid zero-value pixels, preventing sparse-data artifacts from contaminating the representation. This ensures that the extracted features are intrinsically geometry-aligned prior to integration into any downstream multi-modal pipeline.

\section{Problem Analysis}
\label{sec:problem}

While multi-modal 3D detectors achieve state-of-the-art performance, their reliance on bulky visual backbones introduces a critical yet often overlooked inefficiency~\cite{liu2022bevfusion,chen2022autoalignv2}. Standard frameworks typically deploy large 2D-pretrained networks (e.g., Swin Transformers~\cite{liu2021swin} or deep ResNets~\cite{he2016deep}). For instance, in the established BEVFusion framework~\cite{liang2022bevfusion}, the CBSwin-T visual backbone consumes 78.1M parameters, which accounts for 86.6\% of the 90.2M total model capacity. Furthermore, a deep-rooted structural inconsistency exists: the semantic features optimized for 2D classification or detection are intrinsically misaligned with the geometry-aware BEV representations demanded by 3D localization. 
Consequently, these over-parameterized backbones merely act as heavy, inefficient parameter initializers whose capacity for geometry-aware feature extraction remains largely underutilized.

To mitigate this misalignment, an intuitive strategy is to replace 2D semantic pretraining with LiDAR-guided depth completion~\cite{cheng2019learning,wang2023lrru}. However, altering the pretraining objective alone does not overcome the inherent structural limitations of standard network operators. Specifically, conventional convolutional layers operate with spatial invariance, applying identical local aggregation rules across the entire feature map. When processing highly irregular and sparse LiDAR projections, these standard kernels indiscriminately blend valid geometric depth cues with the surrounding empty pixels. Consequently, as the network deepens, the sparse valid signals are rapidly diluted, leading to significant feature degradation within the receptive field. This mechanistic analysis indicates that bridging the semantic-geometric gap demands more than a task-level adjustment; it inherently requires a purpose-built, sparsity-aware architecture.

This insight motivates the design of DeGuNet. 
Rather than relying on large-scale 2D backbones or naively applying standard lightweight convolutions to sparse data, we propose a purpose-built, streamlined architecture. 
By incorporating MPIR blocks~\cite{liu2018partialinpainting} and progressive Guide modules, DeGuNet selectively processes valid geometric signals while isolating invalid regions. 
This targeted, sparsity-aware design allows a compact backbone to inherently align image features with 3D LiDAR geometry, addressing both the parameter redundancy and the structural misalignment.

\section{Methodology}
\label{sec:method}

Our primary objective is to bridge the structural misalignment between standard 2D-pretrained image features and BEV representations demanded by multi-modal 3D object detection. To this end, we propose the DeGuNet, a lightweight, plug-and-play image backbone. Unlike conventional backbones that passively inherit fixed 2D semantic priors, DeGuNet is deliberately pretrained on a LiDAR-guided depth completion task. This specific pretraining ensures that the extracted image features are depth-aware and better aligned with the 3D space prior to any cross-modal fusion. In this section, we first detail the overall macro-architecture and its two-stage training lifecycle. Subsequently, we elaborate on the geometry-guided pretraining paradigm and the core sparsity-aware micro-components designed to mitigate the contamination from invalid projection regions.

\subsection{Overall Architecture}
\label{subsec:overall}

\begin{figure}[t]
  \centering
  \includegraphics[width=\linewidth]{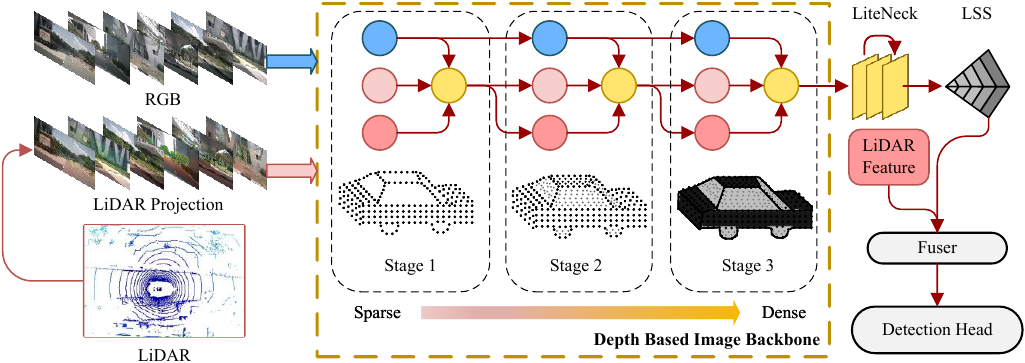}
  \caption{Overview of the proposed detection framework incorporating DeGuNet as a drop-in module. The architecture features a dual-stream design. The depth-based image backbone processes multi-view RGB images and sparse LiDAR projections through progressive stages, generating densified, geometry-aligned features. These features are subsequently decoupled via a LiteNeck and projected into the BEV space for fusion with LiDAR features.}
  \label{fig:overall}
\end{figure}

As illustrated in Fig.~\ref{fig:overall}, the multi-modal detection pipeline is constructed as a dual-stream architecture. It comprises a primary LiDAR processing stream, which can utilize any standard voxel-based~\cite{chen2023voxelnext,shi2022pillarnet,lang2019pointpillars,yan2018second,he2022voxel,zhou2018voxelnet} or point-based encoder~\cite{qi2017pointnet,qi2017pointnet++,shi2020pv,yang20203dssd,feng2024interp,shi2020point}, and a complementary image stream featuring the proposed DeGuNet as its backbone. Operating as a general-purpose feature extractor, DeGuNet takes multi-view RGB images and sparse LiDAR depth projections as parallel inputs. Through progressive stages of cross-modal guidance, it outputs a densified, depth-aware feature representation. 

To ensure generalizability across different LSS-based (Lift-Splat-Shoot) frameworks, the deployment of DeGuNet follows a two-phase lifecycle:
\begin{itemize}
    \item \textbf{Phase 1: Geometry-Guided Pretraining.} An encoder-decoder architecture is constructed to perform depth completion. The encoder (DeGuNet) learns to extract geometry-aligned features from dense RGB and sparse LiDAR inputs, while the decoder upsamples these features solely for depth loss computation.
    \item \textbf{Phase 2: End-to-End Detection Integration.} Upon completion of pretraining, the depth decoder is discarded, while the pretrained encoder is retained and integrated as a drop-in image backbone within the multi-modal 3D detection framework. The features from both sensor streams are then projected into a unified BEV grid and fused for final 3D bounding box prediction. 
\end{itemize}

This decoupled lifecycle guarantees that the image features entering the BEV projection module are aligned with the LiDAR geometry, mitigating the semantic-geometric gap at its source.

\subsection{Geometry-Guided Pretraining Paradigm}
\label{subsec:pretrain}

\begin{figure}[t]
  \centering
  \includegraphics[width=\linewidth]{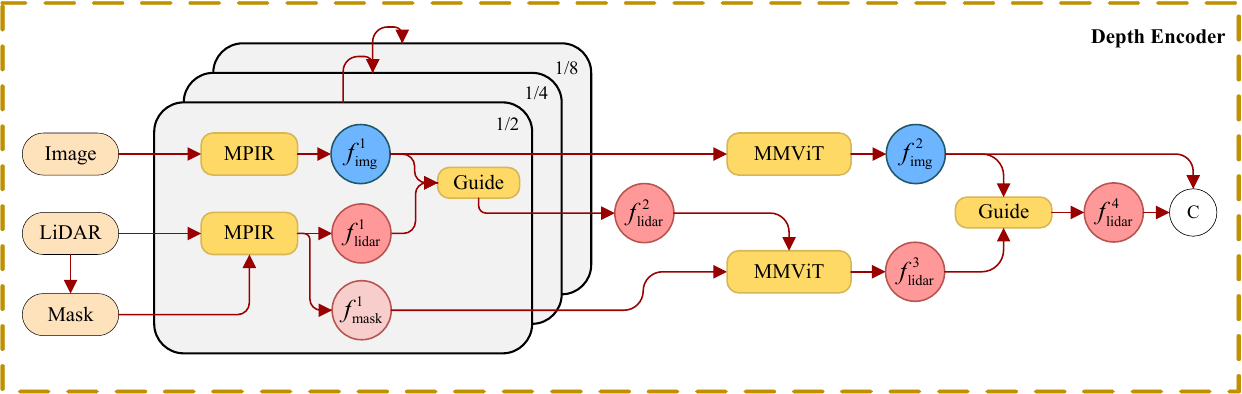}
  \caption{Architecture of the geometry-guided pretraining encoder. The network processes dense multi-view images and sparse LiDAR projections in parallel across three progressive scales. Mask-aware operations ensure that only valid geometric cues guide the representation learning.}
  \label{fig:pretrain_encoder}
\end{figure}

To bridge the spatial awareness gap of 2D-pretrained backbones, we shift the pretraining objective to LiDAR-guided depth completion, forcing the network to directly map 2D textures to 3D geometry. During this phase, the architecture functions as an encoder-decoder network. As shown in Fig.~\ref{fig:pretrain_encoder}, the encoder (the core DeGuNet) processes multi-camera RGB images and sparse LiDAR projections through three progressive downsampling stages (at $1/2$, $1/4$, and $1/8$ resolutions). To handle the inherent sparsity of LiDAR, the encoder employs specialized mask-aware components to prevent invalid projection regions from corrupting the learned features.

\begin{figure}[t]
    \centering
    \includegraphics[width=0.65\linewidth]{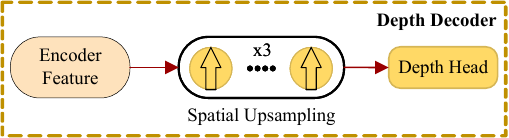}
    \caption{The Depth Decoder, utilized exclusively during the pretraining phase to compute the depth completion loss.}
    \label{fig:decoder}
\end{figure}

A lightweight spatial decoder (Fig.~\ref{fig:decoder}) then upsamples these multi-scale features to predict dense depth maps. Given the ground-truth depth $Z$ and a validity mask $M = \mathbb{I}[Z>0]$, the pretraining is supervised exclusively on valid LiDAR-projected pixels via a multi-scale loss:
\begin{equation}
    \mathcal{L}_{dc} = \sum_{i=1}^{K} \gamma^{K-i} \left( 0.5 \| (\hat{Z}^{(i)} - Z) \odot M \|_{1} + 0.5 \| (\hat{Z}^{(i)} - Z) \odot M \|_{2}^{2} \right),
\end{equation}
where $\{\hat{Z}^{(i)}\}_{i=1}^{K}$ are multi-scale depth predictions, $\gamma$ is a scale-weighting factor, and $\odot$ denotes element-wise multiplication. Crucially, this decoder is purely auxiliary. Upon completing the pretraining, it is entirely discarded. The encoder, now possessing robust geometry-aware feature extraction capabilities, will be seamlessly transferred to the downstream 3D detector, providing aligned features without incurring any auxiliary depth prediction overhead during inference.

\begin{figure}[t]
    \centering
    \includegraphics[width=0.65\linewidth]{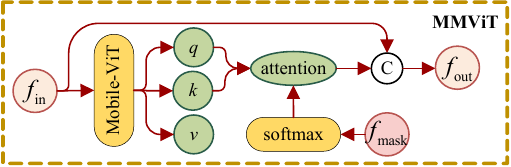}
    \caption{Architecture of the MMViT block, which applies mask-aware attention to prevent invalid depth from polluting global context.}
    \label{fig:mmvit}
\end{figure}

\begin{figure}[t]
    \centering
    \includegraphics[width=0.74\linewidth]{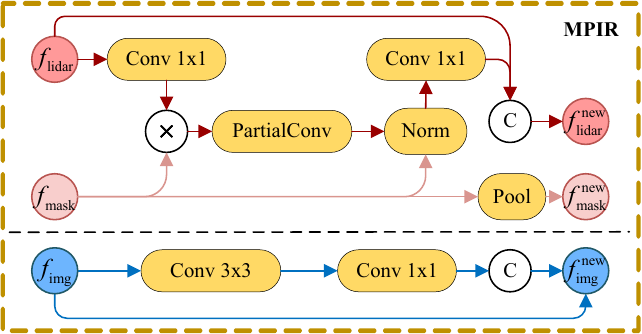}
    \caption{Architecture of the MPIR. It employs mask-guided partial convolutions on the LiDAR stream to prevent invalid region contamination during local feature extraction.}
    \label{fig:mpir}
\end{figure}

\subsection{Sparsity-Aware Feature Extraction}
\label{subsec:sparsity}

Standard convolutional layers apply spatial filters uniformly across their receptive fields. When applied to unstructured LiDAR projections, which predominantly consist of empty pixels, standard kernels inevitably aggregate zeros~\cite{uhrig2017sparsity, graham20183d}. This leads to significant feature degradation and the dilution of valid geometric cues. To prevent this invalid-region degradation, DeGuNet employs a dedicated sparsity-aware design throughout both its local and global feature extraction stages.

To capture long-range spatial dependencies at lower resolutions, DeGuNet utilizes the Masked Mobile Vision Transformer (MMViT) block, as shown in Fig.~\ref{fig:mmvit}. Standard self-attention mechanisms compute global token interactions, which would erroneously allow invalid background nodes to influence the representations of valid geometric structures. To avert this, MMViT introduces a mask-aware attention mechanism~\cite{mao2021voxel,fan2022embracing}. The binary mask $f_{\text{mask}}$ is dynamically downsampled, reshaped, and injected into the attention matrix calculation. By masking out invalid tokens with a large negative value prior to the softmax normalization, MMViT mathematically guarantees that the attention weights for empty regions approach zero. Consequently, the global context aggregation is confined exclusively to valid geometric nodes, preserving the absolute integrity of the spatial representations.

Complementary to the global context aggregation at deeper stages, local feature extraction is handled by the MPIR block, as illustrated in Fig.~\ref{fig:mpir}. The MPIR block processes the multi-modal inputs through a decoupled dual-branch architecture. The image branch processes dense RGB features ($f_{\text{img}}$) using standard depth-wise separable convolutions to capture continuous semantic textures. Conversely, the sparse LiDAR branch ($f_{\text{lidar}}$) incorporates mask-guided partial convolutions~\cite{liu2018partialinpainting}. Given the binary mask $f_{\text{mask}}$ indicating the locations of valid depth values, the partial convolution restricts its weight updates and feature aggregations to these valid pixels. A subsequent pooling operation dynamically updates $f_{\text{mask}}$ for the next layer. This mechanism ensures that the geometric representation remains uncontaminated by the void regions, even as the receptive field expands in deeper layers.

\subsection{Progressive Cross-Modal Guidance and Integration}
\label{subsec:guide}

While the MPIR and MMViT blocks ensure the purity of geometric features by preventing invalid-region pollution, relying solely on sparse LiDAR projections limits the representational density of the network. To address this, DeGuNet incorporates a progressive cross-modal guidance mechanism, operating at the early network stages (e.g., $1/2$ and $1/4$ input resolutions). This early fusion strategy injects high-resolution, dense RGB semantics into the sparse geometric stream, effectively densifying the geometry-aware representations.

To ensure strict alignment between geometric validity and feature resolution throughout the guidance process, the binary mask is recursively updated during downsampling via a max-pooling operation:
\begin{equation}
     f_{\text{mask}}^{(l+1)} = \text{MaxPool}_{s \times s}(f_{\text{mask}}^{(l)}),
\end{equation}
where $f_{\text{mask}}^{(l)}$ denotes the binary validity mask at layer $l$, $f_{\text{mask}}^{(0)} = M$ represents the initial valid mask derived from the raw LiDAR projection, and $s$ is the downsampling stride.

\begin{figure}[t]
    \centering
    \includegraphics[width=0.55\linewidth]{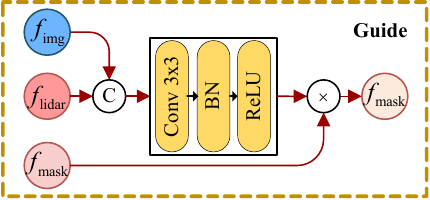}
    \caption{Architecture of the progressive Guide module. It injects dense image semantics into the sparse LiDAR stream.}
    \label{fig:guide}
\end{figure}

Specifically, we design a lightweight Guide module (illustrated in Fig.~\ref{fig:guide}) to perform this cross-modal injection. Let $f_{\text{img}}$ and $f_{\text{lidar}}$ denote the extracted image and LiDAR features at a given scale, respectively. The Guide module concatenates these multi-modal features and processes them through a standard convolutional block. Crucially, to prevent the dense image semantics from erroneously diffusing into the invalid geometric regions, the fused output is tightly bounded by the geometric mask $f_{\text{mask}}$. The operation is formulated as:
\begin{equation}
    f_{\text{out}} = \sigma(\text{BN}(\text{Conv}([f_{\text{img}}, f_{\text{lidar}}]))) \odot f_{\text{mask}},
\end{equation}
where $[\cdot, \cdot]$ denotes channel-wise concatenation, $\text{Conv}$ represents a $3 \times 3$ convolution, $\text{BN}$ is Batch Normalization, $\sigma$ is the ReLU activation function, and $\odot$ denotes element-wise multiplication. By enforcing this mask multiplication, the Guide module ensures that semantic injection is rigorously confined to valid geometric pixels.

Finally, to enable plug-and-play compatibility with various LSS-based architectures, the densified features are processed by a LiteNeck. It structurally decouples the representation into Depth Logits for categorical depth distribution and Context Features for downstream 3D detection. This design allows DeGuNet to be directly integrated into standard Lift-Splat-Shoot modules without requiring any architectural modifications to the baseline.

\section{Experiments}
\label{sec:exp}

\subsection{Experimental Setup}
\label{subsec:setup}

\noindent\textbf{Dataset and Evaluation Metrics.} We evaluate our method on the nuScenes dataset~\cite{caesar2020nuscenes}. It contains 1,000 driving scenes divided into 28,130 training and 6,019 validation samples, providing 360$^\circ$ LiDAR points and synchronized 6-camera images. Standard evaluation metrics include mean Average Precision (mAP) and nuScenes Detection Score (NDS). To comprehensively assess the efficacy of the proposed lightweight module, we also evaluate system-level metrics, including the total parameter count and inference Frames Per Second (FPS).

\noindent\textbf{Implementation Details.} The framework is implemented based on MMDetection3D~\cite{mmdet3d2020}. Point clouds are voxelized within [-54m, 54m] for the X and Y axes, and [-5m, 3m] for the Z axis, with a grid resolution of $0.075\text{ m} \times 0.075\text{ m} \times 0.2\text{ m}$. Multi-view input images are resized to a resolution of 256$\times$704. DeGuNet is then integrated as a drop-in replacement for the original image backbone in the selected frameworks, keeping the subsequent fusion and detection heads unchanged.

For all baseline comparisons, we follow the official open-source implementations and their official configurations, including pretraining recipes, learning rates, optimizers, and training schedules. For inference speed evaluation, FPS is measured on a single NVIDIA A100 GPU with batch size 1, using each method's recommended best-performing inference setup. 

\subsection{The Superiority of Depth-Guided Architecture}
\label{subsec:superiority}

\begin{table}[t]
    \centering
    \caption{Comparison of different backbones pretrained on the KITTI depth dataset~\cite{geiger2012we} and fine-tuned for downstream 3D detection.}
    \label{tab:table1}
    
    \renewcommand{\arraystretch}{1.25} 
    
    \resizebox{0.70\textwidth}{!}{%
    
        \begin{tabular}{l *{3}{w{c}{3.1cm}}} 
        \toprule
        \textbf{Backbone} & \textbf{Depth RMSE (m)} $\downarrow$ & \textbf{3D Det (mAP)} $\uparrow$ & \textbf{Img Params} $\downarrow$ \\
        & \textit{(Pre-train Task)} & \textit{(Downstream)} & \\
        \midrule
        ResNet         & 3.44 & 62.11 & \underline{$\sim$ 0.32M} \\
        MobileViT-XS   & 2.54 & 63.27 & $\sim$ 2.36M \\
        MobileViT-XXS  & 3.71 & 62.09 & $\sim$ 1.33M \\
        Swin-T         & \underline{1.83} & \underline{64.01} & $\sim$ 31.82M \\
        \textbf{DeGuNet (Ours)} & \textbf{0.74} & \textbf{69.40} & \textbf{$\sim$ 0.31M} \\
        \bottomrule
        \end{tabular}%
    }
\end{table}

\begin{table}[t]
    \centering
    \caption{Comparison between standard Auxiliary Depth training strategies and the proposed DeGuNet across different baselines.}
    \label{tab:Baseline}
        \renewcommand{\arraystretch}{1.25} 
        \resizebox{0.65\textwidth}{!}{%
        \begin{tabular}{l *{3}{w{c}{2.2cm}}}
        \toprule
        \textbf{Framework} & \textbf{mAP} $\uparrow$  & \textbf{FPS} $\uparrow$ & \textbf{Params} $\downarrow$ \\
        \midrule
        BEVFusion~\cite{liang2022bevfusion}+ BEVDepth    & 64.8 & 1.1 & 20.7M \\
        BEVFusion~\cite{liu2022bevfusion} + BEVDepth   & 62.8 & 4.2 & 19.6M \\
        GraphBEV~\cite{song2024graphbev}  + BEVDepth        & 68.7 & 4.7 & 12.7M \\
        \midrule
        \rowcolor{gray!10}
        BEVFusion~\cite{liang2022bevfusion} + \textbf{Ours}  & \underline{70.3} & 4.8 & 15.3M \\
        \rowcolor{gray!10}
        BEVFusion~\cite{liu2022bevfusion} + \textbf{Ours} & 69.4 & \textbf{5.1} & \textbf{11.1M} \\
        \rowcolor{gray!10}
        GraphBEV~\cite{song2024graphbev}  + \textbf{Ours}      & \textbf{70.4} & \underline{5.0} & \underline{12.2M} \\
        \bottomrule
        \end{tabular}%
    }
\end{table}

\noindent\textbf{The Necessity of Sparsity-Aware Architecture.} A seemingly intuitive approach to bridge the modality gap is to pre-train standard networks on a depth completion task. However, empirical results in Table~\ref{tab:table1} reveal the inherent limitations of this naive substitution. When pretrained on KITTI depth data, standard convolutional networks and lightweight transformers with comparable parameter counts (e.g., ResNet and MobileViT-XXS) exhibit high Root Mean Square Errors (3.44 m and 3.71 m, respectively) and yield limited downstream 3D detection mAP (62.11 and 62.09). Even when scaling up the parameter count significantly---such as with Swin-T (31.82M parameters)---the geometric alignment remains sub-optimal, yielding an RMSE of 1.83 m and an mAP of 64.01. This indicates that standard spatial operators are ill-equipped to process unstructured LiDAR projections without extensive zero-value contamination. In contrast, DeGuNet, incorporating dedicated MPIR and MMViT blocks, prevents invalid-region pollution. With merely 0.31M parameters, it achieves a substantially lower RMSE of 0.74 m and a superior downstream mAP of 69.40, proving that a structurally sparsity-aware design is a prerequisite for effective geometric alignment.

\noindent\textbf{Plug-and-Play Generalization vs. Auxiliary Depth.} Rather than shifting the pretraining objective, many existing frameworks employ an auxiliary depth paradigm (e.g., BEVDepth), which introduces additional depth supervision during the downstream detector training. As shown in Table~\ref{tab:Baseline}, forcibly appending auxiliary depth losses to existing frameworks yields suboptimal gains and leaves the system burdened by the heavyweight 2D image backbone. For instance, BEVFusion~\cite{liang2022bevfusion} with BEVDepth only reaches 64.8 mAP at 1.1 FPS. Conversely, replacing the original backbone entirely with the pretrained DeGuNet functions establishes a direct, parameter-efficient geometric injection mechanism.  Across established baselines (BEVFusion and GraphBEV), integrating DeGuNet consistently elevates the mAP beyond 69.0 while accelerating the inference speed and reducing the overall parameter footprint (e.g., dropping GraphBEV's total parameters to 12.2M).

\begin{table*}[htbp]
    \centering
    \caption{Performance comparison on the nuScenes validation set. Detailed definitions for Parameter Efficiency (Eff.) and Gain are provided in Sec.~\ref{subsec:main_results}. All performance gains are computed against each method's own LiDAR-only baseline implementation.}
    \label{tab:main_results}
    
    \renewcommand{\arraystretch}{1.25} 
    
    \resizebox{\textwidth}{!}{%
    
        \begin{tabular}{l *{7}{w{c}{2.35cm}}}
        \toprule
        \textbf{Method} & 
        \begin{tabular}{@{}c@{}}\textbf{Img} \\ \textbf{Backbone}\end{tabular} & 
        \begin{tabular}{@{}c@{}}\textbf{Img} \\ \textbf{Params} $\downarrow$\end{tabular} & 
        \begin{tabular}{@{}c@{}}\textbf{Total} \\ \textbf{Params} $\downarrow$\end{tabular} & 
        \begin{tabular}{@{}c@{}}\textbf{Img} \\ \textbf{Param Ratio} $\downarrow$\end{tabular} & 
        \textbf{mAP} $\uparrow$ & \textbf{NDS} $\uparrow$ & \textbf{Eff. / Gain} $\uparrow$ \\
        \midrule
        BEVFusion'22~\cite{liang2022bevfusion}                 & CBSwin-T      & 78.1M & 90.2M & 86.61\% & 69.60 & 72.1  & 0.06 / 4.70 \\
        DeepInteraction++'25~\cite{yang2025deepinteraction++}  & Swin-T        & 33.2M & 62.8M & 52.95\% & 70.63 & \textbf{73.3}  & 0.08 / 2.60 \\
        EA-LSS'23~\cite{hu2023ealss} & CBSwin-T & 78.33M & 153.7M & 50.96\% & 70.90 & 72.8  & 0.08 / 6.00 \\
        SparseFusion'23~\cite{xie2023sparsefusion}             & Swin-T        & 30.3M & 43.8M & 69.18\% & 70.40 & 72.8  & 0.09 / 2.70 \\
        FocalFormer3D'23~\cite{chen2024focalformer3d}          & InternImage   & 44.3M & 65.0M & 68.15\% & 70.50 & \underline{73.1}  & 0.09 / 4.10 \\
        UVTR'22~\cite{li2022unifying}                          & R101          & 47.7M & 88.8M & 53.70\% & 65.40 & 70.2  & 0.09 / 4.60 \\
        BEVFusion'23~\cite{liu2022bevfusion}                   & Swin-T        & 31.8M & 40.8M & 77.94\% & 68.50 & 71.4  & 0.12 / 3.84 \\
        CMT'23~\cite{yan2023cross}                             & VoVNet        & 70.6M & 86.7M & 81.41\% & 70.30 & 72.9  & 0.12 / \textbf{8.16} \\
        ObjectFusion'23~\cite{cai2023objectfusion}             & Swin-T        & 31.8M & 43.9M & 72.43\% & 69.80 & 72.3  & 0.14 / 4.60 \\
        DeepInteraction'22~\cite{yang2022deepinteraction}      & Swin-T        & 28.7M & 57.9M & 49.61\% & 69.85 & 72.7  & 0.17 / 4.80 \\
        GraphBEV'24~\cite{song2024graphbev}                    & Swin-T        & 31.8M & 42.8M & 74.30\% & 70.10 & 72.9  & 0.17 / 5.44 \\
        IS-Fusion'24~\cite{yin2024fusion}                      & Swin-T        & 29.1M & 48.8M & 59.63\% & 71.00 & 72.7  & 0.20 / 5.90 \\
        UniTR'23~\cite{wang2023unitr}                          & Shrunk ViT    & 9.4M  & \underline{12.2M} & 76.49\% & 70.00 & \underline{73.1}  & 0.29 / 3.60 \\
        AutoAlignV2'22~\cite{chen2022autoalignv2}              & CSPNet~\cite{2020wangcsp} & 4.0M & 13.2M & 30.30\% & 67.10 & 71.2  & 0.92 / 3.69 \\

\midrule 
        \rowcolor{gray!10}
        BEVFusion'22~\cite{liang2022bevfusion}  + \textbf{Ours} & DeGuNet      & 1.11M & 15.3M ({\color{upgreen}\textbf{-74.9M}}) & 7.25\%  & 70.30 ({\color{upgreen}\textbf{+0.70}}) & 72.5 ({\color{upgreen}\textbf{+0.4}}) & 4.86 / 5.40 \\
        \rowcolor{gray!10}
        EA-LSS'23~\cite{hu2023ealss} + \textbf{Ours} & DeGuNet & 17.0M & 92.1M ({\color{upgreen}\textbf{-61.6M}}) & 18.46\% & \underline{71.10} ({\color{upgreen}\textbf{+0.20}}) & 72.9 ({\color{upgreen}\textbf{+0.1}}) & 0.36 / 6.20 \\
        \rowcolor{gray!10}
        GraphBEV'24~\cite{song2024graphbev}     + \textbf{Ours} & DeGuNet      & \underline{0.81M} & \underline{12.2M} ({\color{upgreen}\textbf{-30.6M}}) & 6.64\%  & 70.40 ({\color{upgreen}\textbf{+0.30}}) & 72.9 ({\color{upgreen}\textbf{+0.0}}) & 7.09 / 5.74 \\
        \rowcolor{gray!10}
        BEVFusion'23~\cite{liu2022bevfusion}    + \textbf{Ours} & DeGuNet      & \textbf{0.31M} & \textbf{11.1M} ({\color{upgreen}\textbf{-29.7M}}) & \underline{2.79\%}  & 69.40 ({\color{upgreen}\textbf{+0.90}}) & 72.7 ({\color{upgreen}\textbf{+1.3}}) & \underline{15.29} / 4.74 \\
        \rowcolor{gray!10}
        IS-Fusion'24~\cite{yin2024fusion}       + \textbf{Ours} & DeGuNet      & \textbf{0.31M} & 19.7M ({\color{upgreen}\textbf{-29.1M}}) & \textbf{1.57\%} & \textbf{71.30} ({\color{upgreen}\textbf{+0.30}}) & 73.0 ({\color{upgreen}\textbf{+0.3}}) & \textbf{20.00} / 6.20 \\
        \bottomrule
        \end{tabular}%
        
    }
\end{table*}

\subsection{Main Results and System Efficiency}
\label{subsec:main_results}

\noindent\textbf{Performance and Parameter Efficiency.} We evaluate DeGuNet against state-of-the-art multi-modal 3D detectors on the nuScenes validation set (Table~\ref{tab:main_results}). To rigorously quantify the architectural overhead, we introduce Parameter Efficiency (Eff.), defined as the absolute mAP improvement per million image-side parameters. It is calculated as $\frac{\text{Multi-modal mAP} - \text{LiDAR-only mAP}}{\text{Image Parameters (M)}}$. In this context, the absolute improvement over each method's respective baseline is denoted as Gain. Furthermore, the absolute capacity of the image branch (comprising both backbone and neck components) and its proportion relative to the entire system are explicitly reported as Img Params and Img Param Ratio, respectively.

As detailed in Table~\ref{tab:main_results}, conventional frameworks rely on parameter-intensive 2D pretrained backbones consuming roughly 29M to 78M parameters. While these heavy architectures provide absolute mAP gains, their parameter efficiency remains exceedingly low, generally falling below 0.30. This indicates a systematic underutilization of parameters when forcing 2D semantic extractors to align with 3D geometric spaces.

In contrast, substituting standard visual backbones with DeGuNet in established baselines yields consistently higher detection performance while systematically eliminating parameter redundancy. On EA-LSS~\cite{hu2023ealss}, DeGuNet reduces the total/image-side parameters from 153.7M/78.33M to 92.1M/17.0M, while improving mAP/NDS from 70.9/72.8 to 71.1/72.9. On IS-Fusion~\cite{yin2024fusion}, it reduces the total/image-side parameters from 48.8M/29.1M to 19.7M/0.31M, while improving mAP/NDS from 71.0/72.7 to 71.3/73.0. The variation in reported image-side parameters comes from baseline-specific pre-fusion camera modules and the adaptive dimension transformation within LiteNeck, while the core DeGuNet backbone remains strictly at 0.31M parameters. Consequently, DeGuNet achieves positive gains across all tested baselines and reaches an Eff. score of 20.00 on IS-Fusion, validating the effectiveness of depth-guided representations. Beyond parameter reduction, this structural streamlining translates to significant computational gains (Table~\ref{tab:efficiency_analysis}). Compared to the BEVFusion baseline, DeGuNet accelerates inference to 5.1 FPS (a $1.16\times$ speedup) and drastically reduces GPU memory consumption from 20.46GB to 6.86GB (a 66.5\% reduction), proving its superior deployment efficiency.

\begin{table*}[tbp]
    \centering 
    \caption{Computational efficiency comparison on an A100 GPU. The table details DeGuNet's performance in inference speed, memory consumption, and image feature extraction time.}
    \label{tab:efficiency_analysis}
    
    \renewcommand{\arraystretch}{1.25} 
    
    \resizebox{\textwidth}{!}{%
    
        \begin{tabular}{l *{6}{w{c}{2.8cm}}}
        \toprule
        \textbf{Method} & \textbf{FPS} $\uparrow$ & \textbf{Memory} $\downarrow$ & \textbf{Img Backbone} $\downarrow$ & \textbf{Total Img} $\downarrow$ & \textbf{Speedup} $\uparrow$ & \textbf{Memory Reduction} $\uparrow$ \\
        \midrule
        DeepInteraction \cite{yang2022deepinteraction} & 0.2 & 43.11GB  & 1201.6ms & 2162.7ms & 0.045$\times$ & -110.8\% \\
        BEVFusion \cite{liang2022bevfusion}            & 0.3 & 41.01GB  & 1258.1ms & 2236.3ms & 0.068$\times$ & -100.5\% \\
        FUTR3D  \cite{chen2023futr3d}                  & 2.2 & 24.03GB  & 221.8ms  & 221.8ms  & 0.50$\times$  & -17.4\%  \\
        UVTR    \cite{li2022unifying}                  & 2.9 & 23.36GB  & 109.1ms  & 110.7ms  & 0.66$\times$  & -14.2\%  \\
        TransFusion-LC \cite{bai2022transfusion}       & 4.0 & 22.17GB  & \underline{18.04ms}  & \textbf{18.04ms}  & \underline{0.91$\times$}  & \underline{-8.4\%}   \\
        BEVFusion \cite{liu2022bevfusion}              & \underline{4.4} & \underline{20.46GB}  & 21.12ms  & 51.58ms  & 1.0$\times$   & -        \\
        \textbf{DeGuNet (Ours)}                        & \textbf{5.1} & \textbf{6.86GB} & \textbf{7.59ms} & \underline{40.76ms} & \textbf{1.16$\times$} & \textbf{66.5\%} \\
        \bottomrule
        \end{tabular}%
        
    }
\end{table*}

\begin{table}[t]
    \centering
    \caption{Incremental addition of DeGuNet components starting from the LiDAR-only baseline.}
    \label{tab:ablation_incremental}
    
    \renewcommand{\arraystretch}{1.25} 
    \resizebox{0.65\textwidth}{!}{%
    
        \begin{tabular}{l *{3}{w{c}{3.2cm}}}
        \toprule
        \textbf{Components} & \textbf{mAP} $\uparrow$ & \textbf{$\Delta$mAP} $\uparrow$ & \textbf{Params} $\downarrow$ \\
        \midrule
        LiDAR Baseline                   & 64.66 & --     & \textbf{10.30M} \\
        \quad + MPIR blocks              & 66.14 & +1.48  & \underline{10.32M} \\
        \quad + Guide modules            & 67.24 & +2.58  & 10.41M \\
        \quad + MMViT                    & \underline{68.38} & \underline{+3.72}  & 10.61M \\
        \quad + LiteNeck                 & \textbf{69.40} & \textbf{+4.74}  & 11.10M \\
        \bottomrule
        \end{tabular}%
        
    }
\end{table}

\begin{table}[t]

    \centering
    \caption{Component removal analysis demonstrating the performance drop when removing modules from the full DeGuNet.}
    \label{tab:ablation_removal}
    
    \renewcommand{\arraystretch}{1.25} 
    \resizebox{0.65\textwidth}{!}{%
    
        \begin{tabular}{l *{3}{w{c}{3.2cm}}}
        \toprule
        \textbf{Components} & \textbf{mAP} $\uparrow$ & \textbf{$\Delta$mAP} $\uparrow$ & \textbf{Params} $\downarrow$ \\
        \midrule
        Full w/o LiteNeck          & \textbf{68.38} & \textbf{-1.02}  & \textbf{10.61M} \\
        Full w/o MMViT             & 67.24 & -2.16  & \underline{10.95M} \\
        Full w/o Guide             & 66.14 & -3.26  & 11.05M \\
        Full w/o MPIR              & 66.59 & -2.81  & 11.13M \\
        Full (Standard Conv.)      & \underline{68.01} & \underline{-1.39}  & 11.10M \\
        \bottomrule
        \end{tabular}%
        
    }
\end{table}

\subsection{Ablation Studies}
\label{subsec:ablation}

\textbf{Effectiveness of Core Components.} We dissect the individual contributions of DeGuNet's structural components through incremental addition and targeted removal, using the established \textbf{BEVFusion'23}~\cite{liu2022bevfusion} framework as our foundational baseline. As presented in Tables~\ref{tab:ablation_incremental} and \ref{tab:ablation_removal}, starting from the LiDAR-only baseline (64.66 mAP), integrating MPIR blocks yields a 1.48 mAP improvement, validating the necessity of mask-guided partial convolutions in preserving local geometric features from zero-value contamination. The subsequent addition of progressive Guide modules and MMViT blocks further increases mAP to 67.24 and 68.38, respectively. This confirms that both dense semantic injection and mask-aware global attention are critical for robust representation learning. Finally, the LiteNeck module decouples the output dimensions for LSS projection, finalizing the mAP at 69.40. The removal analysis corroborates these findings; systematically stripping any sparsity-aware module incurs significant performance degradation. Notably, reverting the entire backbone to standard convolutions (Full Standard Conv.) results in a 1.39 mAP drop, reaffirming that standard spatial operators cannot adequately handle sparse geometric projections.

\textbf{Progressive Fusion Strategy.} Table~\ref{tab:progressive_fusion} ablates the cross-modal guidance mechanism across different encoder stages. The baseline configuration (66.14 mAP), which utilizes MPIR blocks but lacks semantic injection, serves as the lower bound. Activating the Guide module at single isolated stages (e.g., 1/2 or 1/4 resolution) provides marginal improvements (+0.44 and +0.75 mAP, respectively). However, employing the progressive fusion strategy across all three hierarchical stages yields the optimal performance of 67.24 mAP (+1.10 over the unguided baseline). This demonstrates that multi-scale, continuous injection of dense high-resolution textures is mathematically essential for maximizing the structural densification of the sparse LiDAR stream.

\section{Conclusion}
\label{sec:Conclusion}
In this work, we address a critical inefficiency and structural misalignment in multi-modal 3D object detection: the heavy reliance on heavyweight, 2D-pretrained visual backbones. Through empirical analysis, we demonstrate that standard 2D semantic priors are intrinsically misaligned with 3D geometry, and that simply shifting to a depth-centric pretraining task is insufficient without a specialized architecture to handle unstructured sparse data. To resolve this, we propose DeGuNet, a modular image backbone explicitly designed for geometry-guided representation learning. By integrating MPIR blocks and progressive Guide modules, DeGuNet fuses multi-view images with sparse LiDAR geometry while rigorously preventing invalid-region contamination. Extensive experiments on the nuScenes dataset demonstrate that integrating DeGuNet into established baselines effectively eliminates parameter redundancy and accelerates inference speed, while simultaneously achieving superior 3D detection accuracy. While efficient, the minimalist design inherently limits its capacity for fine-grained semantic abstraction on rare long-tail categories, presenting a valuable direction for future exploration. We believe this work highlights the necessity of structurally aligned, sparsity-aware feature extraction, offering valuable insights for the future development of multi-modal perception systems.

\begin{table}[tbp]

    \centering
    \caption{Ablation study on our progressive fusion strategy. Fusing features at all three encoder stages yields the best performance.}
    \label{tab:progressive_fusion}
    
    \renewcommand{\arraystretch}{1.25} 
    \resizebox{0.65\textwidth}{!}{%
    
        \begin{tabular}{l *{3}{w{c}{2.8cm}}}
        \toprule
        \textbf{Fusion Stages Active} & \textbf{mAP} $\uparrow$ & \textbf{$\Delta$mAP} $\uparrow$ & \textbf{Params (K)} \\
        \midrule
        None (Baseline)     & 66.14 & --    & 0     \\
        Stage 1/2 only      & 66.58 & +0.44 & +4.7   \\
        Stage 1/4 only      & 66.89 & +0.75 & +18.6  \\
        Stage 1/8 only      & 66.74 & +0.60 & +74.0  \\
        Stages 1/2 + 1/4    & 67.03 & +0.89 & +23.3  \\
        Stages 1/4 + 1/8    & 67.14 & +1.00 & +92.6  \\
        \textbf{All (Ours)} & \textbf{67.24} & \textbf{+1.10} & +97.2  \\
        \bottomrule
        \end{tabular}%
        
    } 
\end{table}

\bibliographystyle{splncs04}
\bibliography{main}

\end{document}